# Extended Breadth-First Search Algorithm

Tamás Kádek[1] and János Pánovics[2]

[1] Department of Computer Science, Faculty of Informatics, University of Debrecen
Debrecen, Hungary
*kadek.tamas@inf.unideb.hu*

[2] Department of Information Technology, Faculty of Informatics, University of Debrecen
Debrecen, Hungary
*panovics.janos@inf.unideb.hu*

**Abstract**
The task of artificial intelligence is to provide representation techniques for describing problems, as well as search algorithms that can be used to answer our questions. A widespread and elaborated model is state-space representation, which, however, has some shortcomings. Classical search algorithms are not applicable in practice when the state space contains even only a few tens of thousands of states. We can give remedy to this problem by defining some kind of heuristic knowledge. In case of classical state-space representation, heuristic must be defined so that it qualifies an arbitrary state based on its "goodness," which is obviously not trivial. In our paper, we introduce an algorithm that gives us the ability to handle huge state spaces and to use a heuristic concept which is easier to embed into search algorithms.

## 1. Introduction

The most basic problem representation technique used by artificial intelligence is state-space representation. However, it can be used to describe only a certain small subset of problems conveniently. For example, even a simple chess puzzle may have too many ways to continue from a particular situation because each piece can move in quite a few directions and potentially more than one square. Of course, we do not have to deal with all the possible moves in a particular situation provided we have a means to mark the cases that are relevant regarding the solution's viewpoint. This can only be done only if we have some additional knowledge about the problem, which is usually represented by a heuristic function. In case we have this additional knowledge, however, it may still be difficult to describe it as a function. For example, in the 8-puzzle game, the heuristic function evaluates a situation as being more appealing if it has more pieces on their correct places, but this measure will fail in some cases.

Before we propose a new algorithm that deals with the above-mentioned problems, we first have to define an extended state-space model, in which it is easier to represent the additional (heuristic) knowledge about the problems. After that, we show an extended breadth-first search (EBFS) algorithm that uses the extended model and is able to handle larger state spaces. Finally, we compare this algorithm with the standard breadth-first search via a particular problem.

## 2. An Extended State-Space Model (ESSM)

Using state-space representation, solutions to problems are obtained by executing a series of well-defined steps. During the execution of each step, newer and newer states are created, which form the state space. States are distinguished from one another based on their relevant properties. Relevant properties are defined by the sets of their possible values, so a state can be represented as an element of the Cartesian product of these sets. Let us denote this Cartesian product by $S$. Possible steps are then operations on the elements of $S$. Let us denote the set of operations by $F$. The state space is often illustrated as a graph, in which nodes represent states, and edges represent operations. This way, searching for a solution to a problem can be done actually using a path-finding algorithm.

We keep the basic idea (i.e., the concepts of states and operations on states) also in the extended state-space model (ESSM). The goal of this generalization is to provide the ability to model as many systems not conforming to the classical interpretation as possible in a uniform manner.

A state-space representation over state space $S$ is defined as a 5-tuple of the form
$$\langle K, \text{initial}, \text{goal}, F, B \rangle,$$
where
- $K$ is a nonempty set containing the initially known states. Of course, $K \subseteq S$. The set of initially known states is usually incomplete, nevertheless, only these states can be used as a starting point to explore the state space by applying the operators. Note that by applying the operators on the elements of *K* an arbitrary number of times, *S* is not necessarily covered.
- *initial* is a Boolean function that selects the initial states from the state space:
$$\text{initial} : S \to \{\text{true, false}\}$$
- *goal* is a Boolean function that selects the goal states from the state space:
$$\text{goal} : S \to \{\text{true, false}\}$$
- $F = \{f_1, f_2, \ldots, f_n\}$ is a set of "forward" functions, which represent the operators in the classical sense. Operators can be used to create a new state (or even a set of new states in the extended model) from a given state.
$$f_i : S \to 2^S$$
- $B = \{b_1, b_2, \ldots, b_m\}$ is a set of "backward" functions, which usually give the states from which a given state can be obtained by applying functions in *F*.
$$b_i : S \to 2^S$$

Some notes:
- The number of initial and goal states is not necessarily known initially, as we may not be able to or may not intend to generate the whole set *S* before or during the search.
- The $n + m = 0$ case is excluded because in that case, nothing would represent the relationship between the states.
- Although the elements of the sets *F* and *B* are formally similar functions, their semantics are quite different. The real set-valued functions in *F* are used to represent nondeterministic operators, while there may be real set-valued functions in set *B* even in case of deterministic operators.

Let us now introduce a couple of concepts:
- *Initial state*: a state *s* for which $s \in S$ and initial(*s*) = true.
- *Goal state*: a state *s* for which $s \in S$ and goal(*s*) = true.
- *Known initial state*: an initial state in *K*.
- *Known goal state*: a goal state in *K*.
- *Edge*: an $\langle s, s', o \rangle \in S \times S \times (F \cup B)$ triple where if $o \in F$, then $s' \in o(s)$, and if $o \in B$, then $s \in o(s')$.
- *Path*: an ordered sequence of edges in the form
$$\langle s_1, s_2, o_1 \rangle, \langle s_2, s_3, o_2 \rangle, \ldots, \langle s_{k-1}, s_k, o_{k-1} \rangle,$$
where $k \geq 2$.

General objective: determine a path from $s_0$ to $s^*$, where $s_0$ is an initial state, and $s^*$ is a goal state.

2.1 A Few Properties of ESSM Representations

For classifying state-space representations, let us define some important properties. Let $p = \langle K, \text{initial}, \text{goal}, F, B \rangle$ a state-space representation over *S*. *p* is said to be
- *deterministic* if for all $s \in S$ and $f \in F$, $|f(s)| \leq 1$. If $|f(s)| = 0$, then we say that the operator represented by the forward function *f* is *not applicable* to state *s*. If for some $s \in S$ and $f \in F$, $|f(s)| > 1$ (i.e., *f* is set-valued), then the representation is called *nondeterministic*. In this case, the operator represented by *f* may generate any state in the result set, even different states on different applications. In this paper, we will only focus on deterministic cases.
- *symmetric* if $\forall s \forall s' (\exists k \ (s' \in f_k(s)) \equiv \exists l \ (s \in b_l(s')))$. This means that for each path *P*, there exists a path *P'* that contains the same state pairs in the same order and contains only functions in *F* or functions in *B*.
- *antisymmetric* if
$$\forall s \forall s' ((\exists k \ (s' \in f_k(s)) \supset \neg \exists l \ (s \in b_l(s'))) \land$$
$$\land (\exists l \ (s \in b_l(s')) \supset \neg \exists k \ (s' \in f_k(s)))).$$
In this case, each edge is given in one way only.
- *strictly symmetric* if
$F = \{f_1, f_2, \ldots, f_n\}$, $B = \{b_1, b_2, \ldots, b_n\}$, and
$$\forall s \forall s' \forall k \ (s' \in f_k(s) \equiv s \in b_k(s')).$$
The definition implies that a strictly symmetric representation is also symmetric.
- *one-way forward* if $B = \varnothing$.
- *one-way backward* if $F = \varnothing$.
- *set up with a single initial state* if there exists one and only one $s_0 \in S$ for which initial($s_0$) = true.
- *set up with multiple initial states* if there exists more than one $s \in S$ for which initial(*s*) = true.

In the extended model, the classical state-space representation is a deterministic, antisymmetric representation set up with a single initial state in the following form:
$$\langle \{s_0\}, s \to (s = s_0), \text{goal}, \{f_1, f_2, \ldots, f_n\}, \varnothing \rangle$$
$$f_i(s) = \begin{cases} \{o_i(s)\} & \text{if } s \in \text{dom}(o_i), \\ \varnothing & \text{otherwise,} \end{cases}$$
where $o_i$ is an operator in the traditional sense, for which $o_i : D \to S$ and $D \subseteq S$ ($i = 1, 2, \ldots, n$).

## 3. Model Restrictions for EBFS

Before describing the EBFS algorithm, we first give the model serving as an adequate representation technique for problems suffering from the above-mentioned drawbacks, i.e., the large number of states and nontrivial heuristic functions. We can now make use of the advantage of

ESSM that more than one state (set *K*) may be defined as the input of the search. The basic idea is that the given states should be relevant. This means that the heuristic function is replaced with the enumeration of the states that are considered (potentially) useful. In other words, the given states are predictably a part of one of the solutions. Similarly to using a heuristic function, this prediction is not necessarily perfect. There is only one limitation: at least one of the given states should be on a path representing a solution. Whenever an initial state is included in *K*, this condition is satisfied.

We can also keep the following properties of the extended model:
- it is allowed to have more than one initial and goal states,
- we are able to use both forward and backward functions, so the representation can be symmetric, antisymmetric, or strictly symmetric.

As mentioned above, we set aside nondeterministic representations for now.

Let us now consider a state-space representation that suffers from the presented problems, i.e., a representation whose state space is big enough and for which it is hard to define heuristic as a function. Such a representation exists for the well-known *n*-queens problem. In this representation, a state is defined by an $n \times n$ Boolean matrix, the cells of which represent the squares of a chessboard. An element of the matrix is true if there is a queen on that square and false if it is empty. We have as many operators as many squares on the chessboard. Note that this representation is far from the best choice when it is about solving this problem. We only chose this because it has the drawbacks described earlier.

## 4. The EBFS Algorithm

The EBFS algorithm extends the BFS algorithm with the ability to run more than one breadth-first search starting from more than one state (the inititally known states). It is particularly useful if the subtrees explored reach one another as illustrated by Figure 1. The dashed line denotes the subtree that is discovered by the standard BFS algorithm starting from $i_1$ if the nearest goal state is $g_1$. However, in case we give also the states $k_1$, $k_2$, $k_3$ besides $i_1$ as potentially useful states, then the discovered part of the graph is smaller, even if $k_1$ did not prove to be useful for finding the solution as the illustration shows.

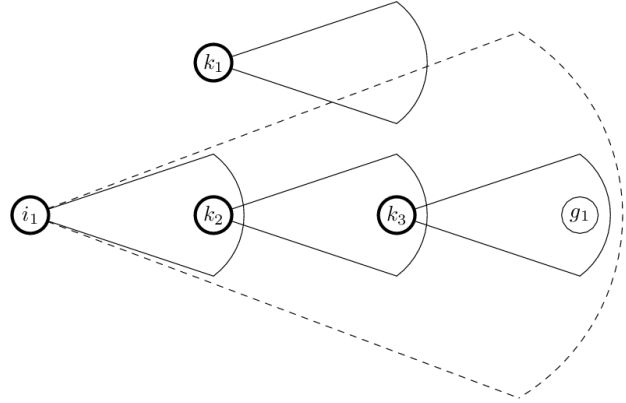

Fig. 1: Subtrees reaching one another.

The EBFS algorithm stores a subgraph of the representation graph during the search. For each node, it stores the state represented by the node as usual. If we have forward functions, we also need to store the forward status (open, closed, or not relevant), forward parents, forward children of the node, as well as the forward distance from each of the initially known states. Note that the forward functions represent the operators in the classical sense. The main difference from BFS at this point is that in case of EBFS, the relationship between the nodes and each initially known state is stored. Because of the ESSM model, we are able to use backward functions as well. If *B* is not an empty set, then we store the above information also for the backward functions. In this case, we need the status of "not relevant". For example, the forward status of a node should be not relevant if it is only discovered using backward functions (because this node is not yet relevant for forward searches). For the sake of simplicity, we now consider *B* an empty set and keep only the B-DISTANCE property so that we can check the termination condition as if we had some backward functions.

### 4.1 The Pseudocode of the Algorithm

```
function NEW-NODE(state)
begin
  STATE[node] ← state
  F-STATUS ← nil
  F-PARENTS[node] ← ∅
  F-CHILDREN[node] ← ∅
  F-DISTANCE[node] ← (∞,∞,…,∞)
  B-DISTANCE[node] ← (∞,∞,…,∞)
  return node
end function
```

```
procedure EBFS
begin
  nodes ← ∅
  i ← 1
  for all k in K do
    new ← NEWNODE(k)
    F-STATUS[new] ← open
    F-DISTANCE[new]_i ← 0
    B-DISTANCE[new]_i ← 0
    nodes ← nodes ∪ new
    i ← i+1
  end for

  while true do
    if { n | n ∈ nodes ∧
      F-STATUS[n] = open } = ∅ then
      terminate unsuccessfully
    end if
    curr ← SELECT(nodes)
    EXPAND(curr, nodes)
    if GOAL-CONDITION(nodes) then
      terminate successfully
    end if
  end while
end procedure
```

The main algorithm is very similar to BFS: it is a series of expansions and termination condition checks.

```
function SELECT(nodes)
begin
  for all n in nodes do
    if n ∈ { m | m ∈ nodes ∧
      min(F-DISTANCE(m)) <=
      min({ min(F-DISTANCE(o)) |
      o ∈ nodes }) } then
      return n
    end if
  end for
  return nil
end function

procedure EXPAND(curr, nodes)
begin
  for all f in F do
    newstate ← f(state(curr))
    node ← SEARCH(nodes, newstate)
    if node = nil or
      F-STATUS[node] = not-relevant then
      if node = nil then
        node ← NEWNODE(newstate)
      end if
      f-status[node] = open
    end if
    F-CHILDREN[curr] ←
      F-CHILDREN[curr] ∪ node
    F-PARENT[node] ← F-PARENT[node] ∪ curr
    F-UPDATE(node, F-DISTANCE[curr])
  end for
end procedure
```

During expansion, we apply all the operators as usual. In the general algorithm, both the forward and backward functions would need to be considered inside the SELECT and EXPAND functions. The SEARCH function checks whether the new state is already in the database.

```
procedure F-UPDATE(node, parent-distance)
begin
  new-distance ← (∞,∞,…,∞)
  for all i in {1, 2, …, count(K)} do
    new-distance_i ← min(F-DISTANCE[node]_i,
      1 + parent-distance_i)
  end for
  if F-DISTANCE[node] <> new-distance then
    F-DISTANCE[node] ← new-distance
    if F-STATUS[node] = closed then
      for all n in F-CHILDREN[node] do
        F-UPDATE(n, new-distance)
      end for
    end if
  end if
end procedure
```

The F-UPDATE function recursively updates the stored information about the nodes whenever an initially known state becomes reachable from another one during the search.

```
function GOAL-CONDITION(nodes)
begin
  for all s in { n | n ∈ nodes ∧
    initial(STATE(n)) } do
    for all g in { n | n ∈ nodes ∧
      goal(STATE(n)) } do
      for all i in {1,2,…,count(K)} do
        if B-DISTANCE(s)_i <> ∞ and
          F-DISTANCE(g)_i <> ∞ then
          return true
        end if
      end for
    end for
  end for
  return false
end function
```

This function checks whether there is an initial state and a goal state such that the goal state can be reached from the initial state via an initially known state. Note that in the simplified case, initial states must also be initially known states.

## 5. Results

The state-space representation described in Section 3 illustrates when the EBFS algorithm can be useful. We ran the EBFS and the classical BFS algorithms with the *n*-queens problem with different values of *n* and summarized the results in the following table:

Table 1: Comparison results

| Problem | BFS | EBFS with 2 known states | EBFS with 3 known states |
|---|---|---|---|
| 5-queens | 453 | 216 | 220 |
| 6-queens | 2 632 | 1 409 | 1 417 |
| 7-queens | 16 831 | 4 434 | 4 439 |
| 8-queens | 118 878 | 46 286 | 46 319 |

The table clearly shows that even with only two initially known states, the number of states explored during the EBFS search until the successful termination is much less than that of the BFS, which is the same as in the case of EBFS with only one initially known state: the initial state (when the board is empty). When we had two initially known states, then for all values of *n*, one of them was the initial state of the problem (which is a sufficient condition for finding a solution if one exists), and the other was a state on a path that represents one of the solutions. The last column shows the case when we added a third state to the two described above with the intention to give a false heuristic: the two states other than the initial state were not reachable from each other. Note that even with including states that are later found to be useless during the search, the number of states explored are still much less than with BFS (of course, this figure highly depends on the selected initially known states).

## 6. Conclusions and Future Work

As you can see from the comparison table, EBFS outperforms the classical BFS algorithm in cases when the state space is large, but we can give a couple of states which we think to form a part of a solution. Introducing the EBFS algorithm was only enabled by creating an extended state-space model first. The EBFS algorithm itself is an extension of the classical BFS algorithm. The question that arises now is how it is possible to extend other graph search algorithms such as uniform-cost search.

**Acknowledgments**

The publication was supported by the TÁMOP-4.2.2.C-11/1/KONV-2012-0001 project. The project has been supported by the European Union, co-financed by the European Social Fund.

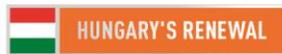
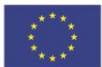

**Tamás Kádek** is an assistant lecturer at the University of Debrecen, Hungary, where he received his master's degree in Computer Science (IT) in 2007. His general research interests include mathematical logic, programming paradigms (imperative, object-oriented, functional, and logic), and artificial intelligence. He has had teaching experience in various fields of IT, including subjects like Logic in Computer Science, Artificial Intelligence, and High-Level Programming Languages.

**János Pánovics** is an assistant lecturer at the University of Debrecen, Hungary, where he received his master's degree in Computer Science (IT) in 1999. His general research interests include programming languages (both low-level and high-level), programming paradigms (imperative, object-oriented, functional, and logic), artificial intelligence, database technologies, and IT education. He has had teaching experience in various fields of IT, including subjects like Assembly Languages, Computer Architectures, High-Level Programming Languages, Data Structures and Algorithms, Database Systems, and Artificial Intelligence.